\documentclass[lnbip]{svmultln}
\usepackage{makeidx}  
\usepackage{times}
\usepackage{helvet}
\usepackage{courier}
\usepackage{amsmath,amssymb,amsfonts,dsfont}
\usepackage{graphicx}
\usepackage{subfigure}
\usepackage{float}
\usepackage{url}
\usepackage{alltt}
\usepackage{xspace}
\usepackage{xcolor}
\usepackage{booktabs}
\usepackage{adjustbox}
\usepackage{wrapfig,fancyvrb}

\usepackage{blindtext,graphicx}
\usepackage[absolute]{textpos}
\begin{document}
\begin{textblock}{16}(3,1)
\noindent\Large Preprint from Proc. of Int. Workshop on BP Innovation with AI (BPAI 2017)
\end{textblock}

\mainmatter              
\title{What Automated Planning can do for \\Business Process Management}
\titlerunning{What Automated Planning can do for BPM}  
%
\author{Andrea Marrella}
\authorrunning{A. Marrella}
%
\tocauthor{Andrea Marrella}
\institute{Sapienza - University of Rome, Italy\\\email{marrella@diag.uniroma1.it}}

\maketitle              

\sloppypar


\begin{abstract}
Business Process Management (BPM) is a central element of today organizations. Despite over the years its main focus has been the support of processes in highly controlled domains, nowadays many domains of interest to the BPM community are characterized by ever-changing requirements, unpredictable environments and increasing amounts of data that influence the execution of process instances. Under such dynamic conditions, BPM systems must increase their level of automation to provide the reactivity and flexibility necessary for process management.
On the other hand, the Artificial Intelligence (AI) community has concentrated its efforts on investigating dynamic domains that involve active control of computational entities and physical devices (e.g., robots, software agents, etc.). In this context, Automated Planning, which is one of the oldest areas in AI, is conceived as a model-based approach to synthesize 
autonomous behaviours in automated way from a model.
In this paper, we discuss how automated planning techniques can be leveraged to enable new levels of automation and support for business processing, and we show some concrete examples of their successful application to the different stages of the BPM life cycle.
\end{abstract}


\newcommand{\LAMA}{\textsc{LAMA}\xspace}
\newcommand{\FASTD}{\textsc{Fast-downward}\xspace}

\newcommand{\Math}[1]{\ensuremath{#1}}
\newcommand{\propername}[1]{\mbox{\textsf{#1}}\xspace}

\newcommand{\myi}{\emph{(i)}\xspace}
\newcommand{\myii}{\emph{(ii)}\xspace}
\newcommand{\myiii}{\emph{(iii)}\xspace}
\newcommand{\myiv}{\emph{(iv)}\xspace}
\newcommand{\myv}{\emph{(v)}\xspace}
\newcommand{\myvi}{\emph{(vi)}\xspace}

\newcommand{\DD}{\mbox{$\cal D$}}                     
\newcommand{\PD}{\mbox{$\cal PD$}}                     
\newcommand{\PR}{\mbox{$\cal PR$}}
\newcommand{\PP}{\mbox{$\cal PP$}}
\newcommand{\CC}{\mbox{$\cal C$}}                     
\newcommand{\RR}{\mathds{R}}
\newcommand{\MK}{\mbox{$\cal K$}}
\newcommand{\NN}{\mathds{N}}
\newcommand{\MB}{\mathds{B}}
\newcommand{\nomove}{\gg}
\newcommand{\Inv}{\textsf{Inv}}
\newcommand{\LAlignMoves}{\Gamma}

\newcommand{\PDDL}[1]{\texttt{#1}}

\def\prparallel{\mathrel{\rangle\!\rangle}}
\def\supparallel{\mathord{|\!|}}
\newcommand{\tuple}[1]{\Math{\langle #1 \rangle}}		

\newcommand{\Pre}{\fontFluents{Pre}}
\newcommand{\Eff}{\fontFluents{Eff}}
\newcommand{\Par}{\fontFluents{Par}}
\newcommand{\Cost}{\fontFluents{Cost}}

\newcommand{\hilight}[1] {\colorbox{yellow}{#1}}

\newcommand{\Prol}[1]{\texttt{\mbox{#1}}}
\newcommand{\kTrue}{\texttt{kTrue}}
\newcommand{\mTrue}{\texttt{mTrue}}

\newcommand{\arrow}[1]{\,\hbox{$\stackrel{\small \mbox{#1}}{\rightarrow}$}\,}
\newcommand{\arrowstar}{\,\hbox{$\stackrel{\mbox{$\ast$}}{\rightarrow}$}\,}

\def\prparallel{\mathrel{\rangle\!\rangle}}
\def\supparallel{\mathord{|\!|}}

\newcommand{\fontFluents}[1]{\mbox{\textit{#1}}}
\newcommand{\fontActions}[1]{\mbox{\textsc{#1}}}

\newcommand{\finishEnvironment}{\hspace*{\fill}\qed}

\section{Introduction}
\label{sec:introduction}

Business Process Management (BPM) is a central element of today organizations due to its potential for increase productivity and saving costs. To this aim,
BPM research reports on techniques and tools to support the design, enactment and optimization of business processes \cite{Aalst2013BPMsurvey}.
Despite over the years the main focus of BPM has been the support of processes in highly controlled domains (e.g., financial and accounting domains), nowadays BPM research is expanding towards new challenging domains (e.g., healthcare \cite{Reichert-healthcare-processes_2007}, smart manufacturing \cite{Seiger2014}, emergency management \cite{Humayoun2009,deLeoni2011}, etc.), characterized by ever-changing requirements, unpredictable environments and increasing amounts of data that influence the execution of process instances \cite{ReichertBook2012}.
Under such dynamic conditions, \emph{BPM is in need of techniques that go beyond hard-coded solutions} that put all the burden on IT professionals, which often lack the needed knowledge to model all possible contingencies at the outset, or this knowledge can become obsolete as process instances are executed and evolve, by making useless their initial effort.
Therefore, there are compelling reasons to introduce \emph{intelligent techniques} that act \emph{autonomously} to provide the reactivity and flexibility necessary for process management  \cite{JODS2015,Hull2016}.

On the other hand, the challenge of building computational entities and physical devices (e.g., robots, software agents, etc.) capable of \emph{autonomous behaviour} under dynamic conditions is at the center of the Artificial Intelligence (AI) research from its origins. At the the core of this challenge lies the \emph{action selection problem}, often referred as the \emph{problem of selecting the action to do next}.
Traditional hard-coded solutions require to consider every option available at every instant in time based on the current context and pre-scripted plans to compute just one next action.
Consequently, they are usually biased and tend to constrain their search in some way.
For AI researchers, the question of action selection is: \emph{what is the best way to constrain this search?} To answer this question, the AI community has tackled the action selection problem through two different approaches \cite{GeffnerBonet:BOOK13-PLANNING}, one based on \emph{learning} and the other based on \emph{modeling}.

In the \emph{learning-based} approach, the controller that prescribes the action to do next is \emph{learned from the experience}. Learning methods, if properly trained on representative datasets, have the greatest promise and potential, as they are able to discover and eventually interpret meaningful patterns for a given task in order to help make more efficient decisions.
For example, learning techniques were recently applied in BPM (see \cite{Maggi2014}) for predicting future states or properties of ongoing executions of a business process.
However, a learned solution is usually a ``black box'', i.e., there is not a clear understanding of how and why it has been chosen. Consequently, the ability to explain why a learned solution has failed and fix a reported quality bug may become a complex task.

Conversely, in the \emph{model-based} approach the controller in charge of action selection is \emph{derived automatically} from a model that expresses the dynamics of the domain of interest, the actions and goal conditions. The key point is that all \emph{models are conceived to be general}, i.e., they are not bound to specific domains or problems. The price for generality is \emph{computational}: The problem of solving a model is computationally intractable in the worst case, even for the simplest models \cite{GeffnerBonet:BOOK13-PLANNING}.

While we acknowledge that both the \emph{learning} and \emph{model-based} approaches to action selection exhibit different merits and limitations, in this paper we focus on a specific model-based approach called \emph{Automated Planning}. Automated planning is the branch of AI that concerns the automated synthesis of autonomous behaviours (in the form of strategies or action sequences) for specific classes of mathematical models represented in compact form. In recent years, the automated planning community has developed a plethora of \emph{planning systems} (also known as \emph{planners}) that embed very effective (i.e., scale up to large problems) domain-independent heuristics, which has been employed to solve collections of challenging problems from several Computer Science domains.

In this paper, we discuss how automated planning techniques can be leveraged for solving real-world problems in BPM that were previously tackled with hard-coded solutions by enabling new levels of automation and support for business processing and we show some concrete examples of their successful application to the different stages of the BPM life cycle.
Specifically, while in Section \ref{sec:basics_planning} we introduce some preliminary notions on automated planning necessary to understand the rest of the paper, in Section \ref{sec:planning_for_BPM} we show how instances of some well-known problems from the BPM literature (such as process modeling, process adaptation
and conformance checking) can be represented as planning problems for which planners can find a correct solution in a finite amount of time.
Finally, in Section \ref{sec:conclusion} we conclude the paper by providing a critical discussion about the general applicability of planning techniques in BPM and tracing future work.

\section{Automated Planning}
\label{sec:basics_planning}

Automated planning addresses the problem of generating autonomous 
behaviours (i.e., \emph{plans}) from a \emph{model} that describes how \emph{actions} work in a \emph{domain} of interest, what is the \emph{initial state} of the world and the \emph{goal} to be achieved. To this aim, automated planning operates on \emph{explicit representations} of states and actions.

The Planning Domain Definition Language (PDDL)~\cite{McDermott:98-PDDL} is a de-facto standard to formulate a compact representation of a \emph{planning problem} $\mathcal{P_R} = \tuple{I,G,\mathcal{P_D}}$, where $I$ is the description of the initial state of the world, $G$ is the desired goal state, and $\mathcal{P_D}$ is the planning domain.
A planning domain $\mathcal{P_D}$ is built from a set of \emph{propositions} describing the \emph{state} of the world in which planning takes place (a state is characterized by the set of propositions that are true) and a set of \emph{actions} $\Omega$ that can be executed in the domain.
An \emph{action schema} $a \in \Omega$ is of the form $a=\tuple{\Par_a,\Pre_a,\Eff_a,\Cost_a}$, where $\Par_a$ is the list of \emph{input parameters} for $a$, $\Pre_a$ defines the \emph{preconditions} under which $a$ can be executed, $\Eff_a$ specifies the \emph{effects} of $a$ on the state of the world and $\Cost_a$ is the \emph{cost} of executing $a$. Both preconditions and effects are stated in terms of the \emph{propositions} in $\mathcal{P_D}$, which can be represented through boolean \emph{predicates}. PDDL includes also the ability of defining \emph{domain objects} and of \emph{typing} the parameters that appear in actions and predicates. 
In a state, only actions whose preconditions are fulfilled can be executed. The values of propositions in $\mathcal{P_D}$ can change as result of the execution of actions, which, in turn, lead $\mathcal{P_D}$ to a new state.
A \emph{planner} that takes in input a planning problem $\mathcal{P_R}$ is able to automatically produce a \emph{plan} $P$, i.e., a controller that specifies which actions are required to transform the initial state $I$ into a state satisfying the goal $G$.

\begin{example} Let us consider a well-known domain in AI: the Blocks World. In Fig. \ref{fig:blocks_world} it is shown an instance of this domain where blocks A, B and C are initial arranged on the table. The goal is to re-arrange the blocks so that C is on A and A is on B.

\begin{figure}[h!]
	\centering
	\includegraphics[width=0.95\columnwidth]{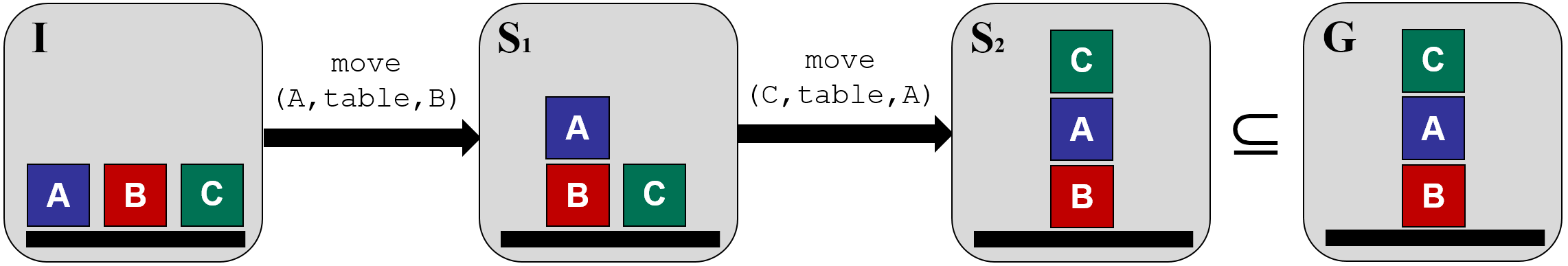}
	\caption{A plan that solves a simple planning problem from the Blocks World domain.}
	\label{fig:blocks_world}
\end{figure}

\noindent
The problem can be easily expressed as a planning problem in PDDL. The variables are the \emph{block locations}: Blocks can be on the table or on top of another block.
Two predicates can be used to express (respectively) that a block is \emph{clear} (i.e., with no block on top) or has another block on \emph{top} of it.
A single planning action \emph{move} is required to represent the movement of a clear block on top of another block or on the table.
Fig.\ref{fig:blocks_world} shows a possible solution to the problem, which consists of first moving A from the table on top of B (state $S_1$), and then on moving C from the table on top of A (state $S_2$). Since $S_2$ is a state satisfying the goal $G$, the solution found is a valid \emph{plan}. Furthermore, if we assume that the cost of any \emph{move} action is equal to 1 (i.e., the cost of the plan corresponds to its length), then the plan found is \emph{optimal}, as it contains the minimum number of planning actions to solve the problem. \qed
\end{example}

There exist several forms of planning models in the AI literature, which result from the application of some orthogonal dimensions \cite{Geffner13}: uncertainty in the initial state (fully or partially known) and in the actions dynamics (deterministic or not), the type of feedback (full, partial or no state feedback) and whether uncertainty is represented by sets of states or probability distributions.
The simplest form of planning where actions are deterministic, the initial state is known and plans are action sequences computed in advance is called \emph{Classical Planning}. For classical planning,
according to \cite{GeffnerBonet:BOOK13-PLANNING}, the general problem of coming up with a plan is NP-hard, since the number of problem states is exponential in the number of problem variables. For example, in a arbitrary Blocks World problem with $n$ blocks, the number of states is exponential in $n$, since the states include all the $n!$ possible towers of $n$ blocks plus additional combinations of lower towers.

Despite its complexity, the field of classical planning has experienced spectacular advances (in terms of scalability) in the last 20 years, leading to a variety of concrete planners that are able to feasibly compute plans with thousands of actions over huge search spaces for real-world problems containing hundreds of propositions. Such progresses have been possible because state-of-the-art classical planners employ powerful \emph{heuristic functions} that are automatically derived by the specific problem and allow to \emph{intelligently} drive the search towards the goal. In addition, since the classical approach of solving planning problems can be too restrictive for environments in which information completeness can not be guaranteed, it is often possible to solve non-classical planning problems using classical planners by means of well-defined transformations \cite{Geffner14}.
A tutorial introduction to planning algorithms and heuristics can be found in \cite{GeffnerBonet:BOOK13-PLANNING}.

\section{Automated Planning for BPM}
\label{sec:planning_for_BPM}

The planning paradigm (in particular in its classical setting) provides a valuable set of theoretical and practical tools to tackle several challenges addressed by the BPM community and its use may lead to several advantages:
\begin{itemize}
\item Planning models are \emph{general}, in the sense that a planner can be fed with the description of any planning problem in PDDL (as defined in Section \ref{sec:basics_planning}) without knowing what the actions and domain stand for, and for any such description it can synthesize a plan achieving the goal. \emph{This means that planners can potentially solve any BPM problem that can be converted into a planning problem in PDDL}.
\item Planning models are \emph{human-comprehensible}, as the PDDL language allows to describe the planning domain and problem of interest in a high-level terminology, which is \emph{readily accessible and understandable by IT professionals}.
\item The \emph{standardized representation} of a planning model in PDDL allows to \emph{exploit a large repertoire of planners and searching algorithms} with very limited effort.
\item Planning models, if encoded with the classical approach, constitute implicit representations of \emph{finite state controllers}, and can be thus \emph{queried by standard verification techniques}, such as Model Checking.
\item BPM environments can invoke planners as \emph{external services}. Therefore, \emph{no expertise of the internal working of the planners} is required to build a plan.
\end{itemize}

A number of research works exist on the use of planning techniques in the context of BPM, covering the various stages of the process life cycle. For the \emph{design-time} phase, existing literature has focused on exploiting planning to automatically generate candidate process models that are able of achieving some business goals starting from a complete or an incomplete description of the process domain. Some research works also exist that use planning to deal with problems for the \emph{run-time} phase, e.g., to adapt running processes to cope with anomalous situations. Finally, for the \emph{diagnosis phase}, the literature reports some works that use planning to perform conformance checking.

In the following sections we discuss in the detail how the use of planning has contributed to tackle the above research challenges from BPM literature.

\subsection{Planning for the Automated Generation of Process Models}
\label{sec:planning_for_modeling}

Process modeling is the first and most important step in the BPM life cycle, which intends to provide a high-level specification of a business process that is independent from implementation and serves as a basis for process automation and verification. Traditional process models are usually well-structured, i.e., they reflect highly repeatable and predictable routine work with low flexibility requirements. All possible options and decisions that can be made during process enactment are statically pre-defined at design-time and embedded in the control-flow of the process. 

\smallskip
\noindent
\textbf{Challenge.}
Current BPM technology is generally based on rigid process models making its application difficult in dynamic and possibly evolving domains, where pre-specifying the entire process model is not always possible.
This problem can be mitigated through specific approaches to process variability~\cite{variabilityACMComp2013}, which allow to customize a base process model by implementing specific variants of the process itself. However, this activity is time-consuming and error-prone when more flexible processes are to be modeled, due to their context-dependent nature that make difficult the specification of all the potential tasks interactions and process variants in advance. \emph{The presence of mechanisms that facilitate the design phase of flexible processes by allowing the automated generation of their underlying models is highly desirable}~\cite{JODS2015}.

\smallskip
\noindent
\textbf{Application of Planning.}
The work~\cite{Schuschel2004} presents the basic idea behind the use of planning techniques for the \emph{automated generation of a process model}.
Process activities can be represented as planning actions together with their preconditions and effects stating what contextual data may constrain the process execution or may be affected after an activity completion. The planning domain is therefore enriched with a set of propositions that characterize the contextual data describing the process domain.
Given an initial description of the process domain, the target is to automatically obtain a plan (i.e., a process model and its control-flow) that consists of process activities contextually selected and ordered in a way to satisfy some business goals.

In the research literature, there are four main approaches that use planning on the basis of the general schema outlined above.
In~\cite{Moreno2007}, the authors exploit 
a planner for modeling processes in SHAMASH, a knowledge-based system for the design of business processes. The planner, which is fed with a semantic representation of the process knowledge, produces a parallel plan of activities that is translated into SHAMASH and presented graphically to the user. The obtained plan proposes the scheduling of parallel activities that may handle time and resource constraints. Notice that the emphasis here is on supporting processes for which one has complete knowledge.

The work~\cite{Ferreira2006} is based on learning activities as planning operators and feeding them to a planner that generates the process model. An interesting result concerns the possibility of producing process models even though the activities may not be accurately described. In such cases, the authors use a best-effort planner that is always able to create a plan, even though the plan may be incorrect. By refining the preconditions and effects of planning actions, the planner will be able to produce several candidate plans, and after a finite number of refinements, the best candidate plan (i.e., the one with the lowest number of unsatisfied preconditions) is translated into a process model.

In the SEMPA approach~\cite{HennebergerHLB08}, process actions are semantically described by specifying their input/output parameters with respect to an ontology implemented in OWL. Starting from such a knowledge, planning is used to derive an action state graph (ASG) consisting of those actions whose execution leads to a given goal from a pre-specified initial state. Then, a process model represented as an UML activity diagram is extracted from the ASG by identifying the required control flow for the process. Interestingly, the planning algorithm implemented in SEMPA provides the ability to build the ASG in presence of initial state uncertainty and with different conflicting goals.

The works~\cite{BPMDS2013,IJSOCA2017} refer to a technique based on partial-order planning algorithms and declarative specifications of process activities in PDDL for synthesizing a library of process templates
to be enacted in contextual scenarios. The resulting templates guarantee that concurrent activities of a process template are effectively independent one from another (i.e., they cannot
affect the same data) and are reusable in a variety of partially specified contextual environments. A key characteristic of this approach is the role of contextual data acting as a driver for process templates generation.

\subsection{Planning for Process Adaptation}
\label{sec:planning_for_enactment}

Process Adaptation is the ability of a process to react to exceptional circumstances (that may or may
not be foreseen) and to adapt/modify its structure accordingly \cite{ReichertBook2012}.
Exceptions can be either anticipated or unanticipated. An anticipated exception can be planned at
design-time and incorporated into the process model, i.e., a (human) process designer can provide an exception handler which is invoked during run-time to cope with the exception. Conversely, unanticipated exceptions generally refer to situations, unplanned at design-time, that may emerge at run-time and can be detected by monitoring discrepancies between the real-world processes and their computerized representation. 

\smallskip
\noindent
\textbf{Challenge.}
In many dynamic application domains (e.g., emergency management, smart manufacturing, etc.), the number of possible anticipated exceptions is often too large, and traditional manual implementation of exception handlers at design-time is not feasible for process designers, who have to anticipate all potential problems and ways to overcome them in advance. Furthermore, anticipated exceptions cover only partially relevant situations, as in such scenarios many unanticipated exceptional circumstances may arise during process execution.
\emph{The management of processes in dynamic domains requires that BPM environments provide real-time monitoring and automated adaptation features during process execution, in order to achieve the overall objectives of the processes still preserving their structure by minimising any human intervention}~\cite{JODS2015}.

\smallskip
\noindent
\textbf{Application of Planning.}
The first work dealing with this research challenge is \cite{Jarvis@AAAIwrk1999}. It discusses how planning can be interleaved with process execution to suggest compensation procedures or the re-execution of activities if some anticipated failure arises. The work \cite{Weske@ADBIS2004} presents an approach for enabling automated process instance change in case of activity failures occurring at run-time that lead to a process goal violation. The approach relies on a partial-order planner for the generation of a new complete process model that complies with the process goal. The generated model is substituted at run-time to the original process that included the failed task.

The above works use planning to tackle anticipated exceptions or to completely redefine the process model when some activity failure arises.
However, in dynamic domains, it would be desirable to adapt a running process by modifying only those parts of the process that need to be changed/adapted and keeps other parts stable, by avoiding to revolutionize the work list of activities assigned to the process participants~\cite{JODS2015}.

In this direction, the SmartPM approach and system \cite{KR2014,ACMTIST2016} provides a planning-based mechanism that requires no predefined handler to build on-the-fly the \emph{recovery procedure} required to adapt a running process instance.
Specifically, adaptation in SmartPM can be seen as reducing the gap between the \emph{expected reality}, i.e., the (idealized) model of reality that reflects the intended outcome of the task execution, and the \emph{physical reality}, i.e., the real world with the actual values of conditions and outcomes.
A recovery procedure is needed during process execution if the two realities are different from each other. A misalignment of the two realities often stems from errors in the tasks outcomes (e.g., incorrect data values) or is the result of exogenous events coming from the environment. If the gap between the expected and physical realities is such that the process instance cannot progress, the SmartPM system invokes an external planner to build a recovery procedure as a plan, which can thereby resolve exceptions that were not designed into the original process.
Notice that a similar framework to tackle process adaptation through planning is also adopted in the research works \cite{BPMDS2011,CollCom2011,CoopIS2012}.

In SmartPM, the problem of automatically synthesize a recovery procedure is encoded as a classical planning problem in PDDL. The planning domain consists of propositions that characterize the contextual data describing the process domain. Planning actions are built from a repository of process activities annotated with preconditions and effects expressed over the process domain. Then, the initial state reflects the physical reality at the time of the failure, while the goal state corresponds to the expected reality. A classical planner fed with such inputs searches for a plan that may turn the physical reality into the expected reality by adapting the faulty process instance.

A similar adaptation strategy is applied in \cite{Bucchiarone@SOCA2011}, which proposes a goal-driven approach for service-based applications to adapt business processes to run-time context changes. Process models include service annotations describing how services contribute to the intended goal. Contextual properties are modeled as state transition systems capturing possible values and possible evolutions in the case of precondition violations or external events. Process and context changes that prevent goal achievement are managed through an adaptation mechanism based on service composition via planning.

Finally, the work \cite{vanBeest2014} proposes a runtime mechanism that uses dependency scopes for identifying critical parts of the processes whose correct execution depends on some shared variables and intervention processes for solving potential inconsistencies between data. Intervention processes are automatically synthesised through a planner based on CSP techniques. While closely related to SmartPM, this work requires specification of a (domain-dependent) adaptation policy, based on volatile variables and when changes to them become relevant.

\subsection{Planning for Conformance Checking}
\label{sec:planning_for_mining}

Within the discipline of process mining, conformance checking is the problem of verifying whether the observed behavior stored in an event log is compliant with the process model that encodes how the process is allowed to be executed to ensure that norms and regulations are not violated. The notion of alignment \cite{process-mining-book} provides a robust approach to conformance checking, which makes it possible to exactly pinpoint the deviations causing nonconformity with a high degree of detail. An alignment between a recorded process execution (log trace) and a process model is a pairwise matching between activities recorded in the log (events) and activities allowed by the model (process activities).

\smallskip
\noindent
\textbf{Challenge.}
In general, a large number of possible alignments exist between a process model and a log trace, since there may exist manifold explanations why a trace is not conforming. It is clear that one is interested in finding the most probable explanation, i.e., one of the alignments with the least expensive deviations (i.e., optimal alignments), according to some function assigning costs to deviations.
The existing techniques to compute optimal alignments against procedural \cite{Adriansyah2013} and declarative \cite{Leoni2012} process models provide ad-hoc implementations of the A* algorithm. The fact is that when process models and event logs are of considerable size the existing approaches do not scale efficiently due to their ad-hoc nature and they are unable to accomplish the alignment task.
\emph{In the era of Big Data, scalable approaches to process mining are desperately necessary, as also advocated by
the IEEE Task Force in Process Mining \cite{process-mining-book}}.

\smallskip
\noindent
\textbf{Application of planning.}
In case of procedural models represented as Petri Nets, the work \cite{ESWA2017} proposes an approach and a tool to encode the original algorithm for trace alignment \cite{Adriansyah2013} as a planning problem in PDDL.
Specifically, starting from a Petri net $N$ and an event log $L$ to be aligned, for each log trace $\sigma_L \in L$ it is built \myi a planning domain $P_D$, which encodes the propositions needed to capture the structure of $N$ and to monitor the evolution of its marking, and three classes of planning actions that represent ``alignment'' moves: synchronous moves (associated with no cost), model moves and log moves; and \myii a planning problem  $P_R$, which includes a number of constants required to properly ground all the domain propositions in $P_D$; in this case, constants will correspond to the place and transition instances involved in $N$. Then, the initial state of $P_R$ is defined to capture the exact structure of the specific log trace $\sigma_L$ of interest and the initial marking of $N$, and the goal condition is encoded to represent the fact that $N$ is in the final marking and $\sigma_L$ has been completely analyzed.
At this point, for any trace of the event log, an external planner is invoked to synthesize a plan to reach the final goal from the initial state, i.e., a sequence of alignment moves (each of which is a planning action) that establish an optimal alignment between $\sigma_L$ and $N$.

Relatively close is the work \cite{DiFrancescomarino15} where authors use planners to recover the missing recording of events in log traces. The concept of missing event recordings is very similar to model moves in \cite{ESWA2017}. However, in \cite{DiFrancescomarino15} it is assumed that all executions are compliant with the model and, hence, every event that is present in the incomplete log trace is assumed to be correct. In other words, they do not foresee log moves.

In case of declarative process models, where relationships among process activities are implicitly defined through logical constraints expressed in the well-known LTL-f (Linear Temporal Logic on finite traces) formalism, the work \cite{AAAI2017} leverages on planning techniques to search for optimal alignments.
A planning domain is encoded to capture the structure of the finite state automata (augmented with special transitions for adding/removing activities to/from a log trace) corresponding to the individual LTL-f constraints that compose the declarative model. The same can be done for the specific trace to be aligned, which is represented as a simple automaton that consists of a sequence of states. In addition, the definition of specific domain propositions allows to monitor the evolution of any automaton.
At this point, the initial state of the planning problem is encoded to capture the exact structure of the trace automaton and of every constraint automaton. This includes the specification of all the existing transitions that connect two different states of the automata. The current state and the accepting states of any trace/constraint automaton are identified as well.
Then, the goal condition is defined as the conjunction of the accepting states of the trace automaton and of all the accepting states of the constraint automata. At this point, a planner is invoked with such inputs to synthesize a plan that establishes an optimal alignment between the declarative model and the log trace of interest

Notably, both the works \cite{ESWA2017} and \cite{AAAI2017} report on results of experiments conducted with several planners fed with combinations of real-life and synthetic event logs and processes. The results show that, when process models and event-log traces are of considerable size, their planning-based approach outperforms the existing approaches based on ad-hoc implementations of the A* algorithm \cite{Adriansyah2013,Leoni2012} even by several orders of magnitude, and they are always able to properly complete the alignment task (while the existing approaches run often out of memory).

\section{Discussion and Conclusion}
\label{sec:conclusion}

We are at the beginning of a profound transformation of BPM due to the recent advances in AI research \cite{Hull2016}. In this context, we have shown how Automated Planning can offer a mature paradigm to introduce autonomous behaviour in BPM for tackling complex challenges in a theoretically grounded and domain-independent way. If BPM problems are converted into planning problems, one can seamlessly update to the recent versions of the best performing automated planners, with evident advantages in term of versatility and customization. In addition, planning systems employ search algorithms driven by intelligent heuristics that allow to scale up efficiently to large problems.

On the other hand, although Planning (in particular in its classical setting) embeds properties desirable for BPM, it imposes some restrictions for addressing more expressive problems, including preferences and non deterministic task effects. Furthermore, planning models require that actions are completely specified in term of I/O data elements, preconditions, and effects, and that the execution context can be captured as part of the planning domain. These aspects can frame the scope of applicability of the planning paradigm to BPM.

It is worth to mention that Automated Planning contributed to tackle challenges also from other Computer Science fields, such as Web Service Composition \cite{Marconi@ICWS2005,Marconi@WWW2005} and Ubiquitous Computing \cite{Georgievski@2016}.
As a future work, we aim at developing a rigorous methodology to acquire relevant literature on the use of planning for BPM and derive an common evaluation framework to systematically review and classify the existing methods.


\small
\bibliographystyle{splncs}
\bibliography{bibliography}

\end{document}